\documentclass[12pt]{article}
\usepackage{graphicx}
\usepackage[utf8]{inputenc} 
\usepackage[T1]{fontenc}    
\usepackage{hyperref}       
\usepackage{url}            
\usepackage{booktabs}       
\usepackage{amsfonts}       
\usepackage{nicefrac}       
\usepackage{xcolor}         

\usepackage{microtype}
\usepackage{graphicx}
\usepackage{subfigure}
\usepackage{amsmath}
\usepackage{amsthm}
\usepackage{mathtools}
\usepackage{tabularx}
\usepackage[numbers,sectionbib]{natbib}
\usepackage[nottoc]{tocbibind}
\usepackage[textsize=tiny]{todonotes}
\usepackage[capitalize,noabbrev]{cleveref}
\usepackage{algorithm}
\usepackage{algpseudocode}
\usepackage{tcolorbox}
\usepackage{bm}
\usepackage[shortlabels]{enumitem}
\usepackage{wrapfig}
\usepackage{amsmath}

\usepackage{tikz}
\usepackage{pgfplots}
\usepackage{environ}
\usepackage{tikz-cd}
\usetikzlibrary{decorations.pathreplacing}

\pgfplotsset{compat=newest}

\AtEndEnvironment{example}{\hfill$\blacksquare$}%
\AtEndEnvironment{definition}{\hfill$\blacksquare$}%

\pgfkeys{/pgfplots/scale/.style={
  x post scale=#1,
  y post scale=#1,
  z post scale=#1}
}

\pgfplotsset{every tick label/.append style={font=\tiny}}

\usetikzlibrary{arrows,backgrounds,decorations.markings,hobby}
\usepgflibrary{shapes.multipart}
\usetikzlibrary{patterns}
\makeatletter
\newcommand{\pgfplotsdrawaxis}{\pgfplots@draw@axis}
\makeatother

\pgfplotsset{only axis on top/.style={axis on top=false, after end axis/.code={
             \pgfplotsset{axis line style=opaque, ticklabel style=opaque, tick style={thick,opaque},
                          grid=none}\pgfplotsdrawaxis}}}
\pgfplotsset{every axis/.style={scale only axis}}
\tikzset{>=latex} 

\tikzstyle{oper}=[rounded corners, draw=black, thick, minimum size = 4mm]
\tikzstyle{input}=[rounded corners, draw=white, thick, minimum size = 4mm]
\tikzstyle{output}=[rounded corners, thick, draw=white, minimum size = 4mm]
\tikzstyle{empty}=[circle, draw=white, minimum size = 4mm]
\makeatletter
\newsavebox{\measure@tikzpicture}
\NewEnviron{scaletikzpicturetowidth}[1]{%
  \def\tikz@width{#1}%
  \begin{lrbox}{\measure@tikzpicture}%
  \BODY
  \end{lrbox}%
  \pgfmathparse{#1/\wd\measure@tikzpicture}%
  \BODY
}
\makeatother
\usepackage{hyperref}

\makeatletter
\renewcommand{\@biblabel}[1]{#1.}
\makeatother

\newcommand{\coNP}{\mathsf{coNP}}
\newcommand{\NP}{\mathsf{NP}}
\newcommand{\p}{\mathsf{P}}

\newcommand{\R}{\mathbb{R}}
\newcommand{\Z}{\mathbb{Z}}

\newcommand{\norm}[1]{||#1||}
\newcommand\inprod[1]{\langle #1 \rangle}
\newcommand{\func}{\circ}
\newcommand{\ACTL}{\text{diag}}
\newcommand{\act}{\sigma}

\newcommand{\dualv}{\eta}

\newcommand{\trace}{\text{tr}}

\newcommand{\relu}{\text{ReLU}}
\newcommand{\obju}{\zeta}
\newcommand{\eig}{\lambda}
\newcommand{\class}{\mathcal{C}}

\DeclareMathOperator*{\argmax}{arg\,max}

\theoremstyle{plain}
\newtheorem{theorem}{Theorem}[section]

\theoremstyle{definition}

\theoremstyle{remark}
\newtheorem{remark}[theorem]{Remark}

\renewcommand{\paragraph}[1]{\vspace{.05in}\noindent\textbf{{#1}.~~}}

\begin{document}
\title{Efficient Symbolic Reasoning for Neural-Network Verification}
\date{}
%
%
\author{\small Zi Wang$^{\dag}$  \;\;   Somesh Jha$^{\dag}$\;\; Krishnamurthy (Dj) Dvijotham$^{\ddag}$ \; \\$^{\dag}$\small University of Wisconsin-Madison 
\; $^{\ddag}$Google Research\\\texttt{\small zw@cs.wisc.edu, jha@cs.wisc.edu, dvij@google.com}  }
\date{}
%
%
%
\maketitle              
\begin{abstract}
The neural network has become an integral part of modern software systems. However, they still suffer from various problems, in particular, vulnerability to adversarial attacks. In this work, we present a novel program reasoning framework for neural-network verification, which we refer to as symbolic reasoning. The key components of our framework are the use of the symbolic domain and the quadratic relation. The symbolic domain has very flexible semantics, and the quadratic relation is quite expressive. They allow us to encode many verification problems for neural networks as quadratic programs. Our scheme then relaxes the quadratic programs to semidefinite programs, which can be efficiently solved.
This framework allows us to verify various neural-network properties under different scenarios, especially those that appear challenging for non-symbolic domains. Moreover, it introduces new representations and perspectives for the verification tasks. We believe that our framework can bring new theoretical insights and practical tools to verification problems for neural networks.  

\end{abstract}
\section{Introduction}

Deep neural networks (DNN) have achieved unprecedented success in many complex tasks and has been integrated into modern software systems~\cite{Krizhevsky_imagenetclassification,nlp_2013}. This brings new challenges and opportunities to the program-verification community to analyze these programs. Like traditional software, DNNs were also shown vulnerable to various attacks. Among them, the most unique and notable are adversarial attacks. 

Adversarial examples were first identified in~\cite{szegedy2014intriguing,goodfellow2015explaining}. Later works found that adversarial attacks are common for different neural-network architectures and tasks~\cite{adv-rnn,adv-metric,adv-transformer}. Adversarial attacks are small perturbations that can cause a prediction to change when they are applied to the inputs, and these perturbations are unnoticeable to humans. This causes serious concerns for DNN safety and reliability because a malicious user can exploit these attacks. Researchers invented various defense mechanisms but they were later shown still vulnerable to those attacks~\cite{CW,papernot2015limitations,madry2018towards}. More recently, the community focuses on certifiable robustness against adversarial attacks~\cite{albarghouthi2021introduction,cav-safety,reluplex,certified_def}.

\paragraph{Neural-network verification} To certify a prediction, we need to estimate the change of the output given an input perturbation. Unfortunately, exact verification has been shown $\NP$/$\coNP$-hard~\cite{jordan2021exactly,reluplex,IUA}. To enable efficient DNN verification, we have to over-approximate the computation of DNNs. As a result, the key challenge is to balance efficiency and precision. 

A large body of verification works is based on the classical abstract interpretation~\cite{abstract_int}. Verification against adversarial attacks requires capturing all possible executions. Abstract interpretation starts by defining an abstract domain that can over-approximate all the inputs, and re-interpret the network execution in terms of the abstract domain~\cite{ai2,symb_int,zonotope}. Because this re-interpretation is sound, and the execution is compositional, abstract interpretation enables a sound verification of the network. However, it remains challenging to verify DNNs with non-linear perturbations and unconventional architectures because the abstract domain is usually defined in terms of linear relations, and reasoning non-linear computations involves many ad-hoc techniques.

\paragraph{Symbolic framework}In this work, we propose a new program-reasoning framework for DNN verification. We refer to this framework as symbolic reasoning~\footnote{Our symbolic reasoning framework can also be viewed as an abstract interpretation example with the symbolic domain.}. 
Our framework contains two distinguished components: symbolic domains and quadratic relations. Symbolic domains have flexible semantics, allowing us to reason tasks that appear challenging for non-symbolic domains. We will elaborate on this paradigm and demonstrate its power for DNN verification. Similar to abstract interpretation, this program reasoning scheme is also compositional, and can be easily adapted to different architectures.

We use quadratic relations to encode the computation and attack constraints rather than the commonly used linear relations. Quadratic encoding is quite expressive. For example, many intractable combinatorial optimization problems can be written as quadratic programs~\cite{maxcut}. As we will demonstrate, it can precisely encode various DNN verification tasks as algebraic formulas. The verification problem is then translated into a quadratic program (QP). This allows us to use mathematical tools to accurately analyze those tasks. Meanwhile, this precise representation also enables us to ask important theoretical questions about the verification problems. We will provide quadratic encodings for various tasks, such as different attacks and activations. Constructing the QPs for different verification tasks amounts to assembling these components together.

Because quadratic encoding is fairly expressive, the resulting QPs are in general hard to solve. To enable efficiency, we consider the semidefinite relaxation of quadratic programs. Semidefinite programming (SDP) is a unique topic where many subjects meet together such as theoretical computer science, optimization, control theory, combinatorics and functional analysis. For example, \cite{flag} introduced flag algebras, which uses SDP to derive bounds for extremal combinatorics problems. In particular, some SDP-induced algorithms are optimal within polynomial time assuming some complexity-theoretical conjectures~\cite{opt_inf2,UGC_inf1_opt}. We believe that formulating the verification problems as semidefinite programs introduces new representations and perspectives for the verification tasks, and brings more theoretical and practical tools to address those problems. 

\paragraph{Relevance} There have been a few works on using SDP to verify DNNs~\cite{lipsdp,certified_def,sdp_rob_local,geolip}. These works appear designed for specific verification tasks and are less accessible to researchers, as some works~\cite{lipopt,sdp_rob_local} claimed that one technique cannot transfer to another setting (see more discussion in~\cref{sec:theory}). We will demonstrate the reasoning framework behind these works to popularize them. Moreover, we show that the paradigm is powerful and can be applied beyond the application scope of these works. In the meantime, there are a few other works aiming to understand the quality and improve the empirical performance of these works~\cite{efficient_sdp,chordal_sparsity,explore_chordal,sdp_quality}. Our paper implies that these works can be more impactful beyond their original scope. We also present a comprehensive discussion of this framework, which sheds light on future directions.

\paragraph{Contributions}To summarize, our paper makes the following contributions:
\begin{enumerate}
    \item We provide a novel systematic neural-network reasoning paradigm and it provides new representations for many verification tasks, i.e., as QPs and SDPs. These representations can bring new theoretical tools, such as matrix analysis and approximation theory, to those tasks (\cref{sec:method,sec:dis});
    \item We present encodings for various mathematical components, which allows us to verify different properties and network structures, especially those that appear challenging for linear-relational non-symbolic domains (\cref{sec:power});
    \item We empirically examine our framework on a specific DNN verification task: $\ell_2$-robustness certification. The evaluation shows that the result from our framework is consistently precise ($60\%$ improvement on average compared to the benchmark), and it can handle practical-size DNNs (\cref{sec:eva}).
\end{enumerate}
\section{Preliminaries}
\paragraph{Notations and definitions}
Let $[n] = \{1,\ldots,n\}$, 
$\R_+ = [0,\infty)$ and $\Z_+$ be the set of positive integers. For a matrix $M\in \R^{m\times n}$, $M^T$ denotes its transpose. For any vector $v\in \R^n$, $\ACTL(v)$ is an $n\times n$ diagonal matrix, with diagonal values $v$. Let $e_n = (1,\ldots, 1)\in \R^n$ be an $n$-dimensional vector of all $1$'s, and $I = \ACTL(e_n)$ is the identity matrix. 
Let $\norm{v}_p$ denote the $\ell_p$ norm of $v$: 
\[\norm{v}_p = \sqrt[\leftroot{-1}\uproot{8}\scriptstyle p]{\sum_{i=1}^n |v_i|^p}.\]
The canonical Euclidean norm is then $\norm{v}_2$, and another commonly considered attack on the input is the $\ell_\infty$-attack:
\[
\norm{v}_\infty = \max_{i\in[n]}{|v_i|}.
\]
Throughout the paper, we consider the $\ell_p$-norm of the input's perturbation for $p\geq 1$.

For a matrix $N$, $\eig(N)$ denotes the eigenvalues of $N$. An $n\times n$ symmetric matrix $M\succeq 0$ means that $M$ is positive semidefinite (PSD). There are three common equivalent conditions for $M\succeq 0$:
\begin{enumerate}
\item All eigenvalues of $M$ are non-negative, i.e., $\eig_{\min}(M)\geq 0$;
\item $x^T M x \geq 0$ for all $x\in \R^n$;
\item $\exists L\in \R^{n\times m}$ such that $ LL^T = M$, where $m\in \Z_+$.
\end{enumerate}
Let $\trace(M)$ be the trace of a square matrix $M$: $\trace(M)=\sum_{i=1}^n M_{ii}.$
The Frobenius inner product of two matrices $A\in \R^{m\times n}$ and $B\in \R^{m\times n}$ is 
\[\inprod{A, B}_F = \trace{(A^TB)}=\sum_{i=1}^m\sum_{j=1}^nA_{ij}B_{ij}.\] A vector function $f: \R^n\rightarrow \R^m$ is an affine transformation if $f(x) = Wx + b$ for $W\in \R^{m\times n}$ and $b\in \R^m$. For two functions $f$ and $g$, $f\func g(x) = f(g(x))$ denotes the composition of $f$ and $g$.

Given two metric spaces $(X, d_X)$ and $(Y, d_Y)$, a function $f: X\rightarrow Y$ is \emph{Lipschitz} continuous if there exists $K > 0$ such that for all $x_1, x_2\in X$,
\begin{equation}\label{eq:lipDef}
    d_Y(f(x_2), f(x_1))\leq Kd_X(x_2, x_1).
\end{equation}
The smallest such $K$ satisfying~\cref{eq:lipDef}, denoted by $K_f$, is called the Lipschitz constant of $f$.

\paragraph{Feed-forward networks}
We start with the standard feed-forward structures.  A neural network $f:\R^m\rightarrow \R^l$ is a composition of affine transformations and activation functions:
\[
f_1(x)=  W^{(1)} x+b^{(1)};\; f_i(x)=  W^{(i)}\act(x)+b^{(i)}, i=2,\ldots,d.
\]
where $W^{(i)}\in \R^{n_{i+1}\times n_{i}}$ is the weight matrix between the layers, $n_1 = m$ and $n_{d+1}=l$, $d$ is the depth of the network, and $b^{(i)}\in \R^{n_{i+1}}$ is the bias term. $\act$, the activation, is an element-wise non-linear function. $f = f_d\func\cdots\func f_1$. 

$f: \R^m\rightarrow \R^l$ has $l$ outputs. Let $f^{(i)}$ be the $i$-th output of $f$. The classification of an input $x$ is $\class(f,x)=\argmax_{i\in[l]}f^{(i)}(x)$. Suppose that the prediction of $x$ is $j$, then $f^{(j)}(x)> f^{(k)}(x)$ for all $k\neq j$. The output of $f$ is called the logit score, and the classification model outputs the class with the highest logit score.

$\Tilde{x}$ is an \emph{adversarial attack} for $x$ if $\norm{\bar{x}-x}_p\leq \epsilon$ and $\class(f,\Tilde{x}) \neq j$.

We use $z$ to denote the output of the $(d-1)$-th layer, the representation layer of the network. Let $w^{(i)}_j$ be the $j$-th row of $W^{(i)}$, then $f^{(i)}(x) = w_j^{(d)}z+b^{(d)}_j$. Verifying whether a prediction changes amounts to maximizing $(w^{(d)}_kz+b^{(d)}_k)-(w^{(d)}_jz+b^{(d)}_j) = (w^{(d)}_k-w^{(d)}_j)z+(b^{(d)}_k-b^{(d)}_j)$. If the maximum value is negative for all $k\neq j$, then this prediction is robust. Therefore, we can use a vector $v$ to denote $w^{(d)}_k-w^{(d)}_j$ and a scalar $c$ to denote $b^{(d)}_k-b^{(d)}_j$. From now on, let us assume $l=1$.

In this paper, we focus on the $\relu$ activation function~\cite{relu}, due to its broad applicability, and the verification literatures often study it~\cite{Baader2020Universal,reluplex,chen2020semialgebraic}. $\relu(x) = \max(x, 0)$ is a piece-wise linear function. \Cref{fig:relu} shows the definition of $\relu$ and its plot. In~\cref{sec:other-act}, we discuss other activation functions than $\relu$. 
    \begin{figure}\normalsize
    

        \centering
        \begin{tikzpicture}
            \begin{axis}[
                axis lines=middle,
                y=1.2cm,
                x=1.2cm,
                xmax=2.5,
                xmin=-2.5,
                xtick={-2,0, 2},
                ymin=-1,
                ymax=2.2,
                ytick={0,1, 2},
                width=2cm
            ]

            \addplot [domain=-10:0, samples=100,
                      thick, red] {0};
                
            \addplot [domain=0:10, samples=100,
            thick, red] {x};

            \node at (axis cs:-0.6,0) [circle, scale=0.3, draw=black!80,fill=black!80] {};
            \node at (axis cs:1.2,1.2) [circle, scale=0.3, draw=black!80,fill=black!80] {};

            \tikzstyle{dashed}=[dash pattern=on 3pt off 3pt,color=blue]
            \draw[dashed] (0,1.2) node[left] {$b$} -- (1.2,1.2) node[right] {$p$};
            \draw[dashed] (1.2,0) node[below] {$a$} -- (1.2,1.2) node[right] {};
            \draw[dashed] (1.2,1.2) node[below] {} -- (-0.6,0) node[right] {};
            \draw[dashed] (0,0) node[below right] {$\Tilde{b}$} -- (-0.6, 0) node[above left] {$\Tilde{p}$};
            \draw[dashed] (-0.6, 0) node[below] {$\Tilde{a}$};
            \addplot[mark=none, dashed, color=blue] coordinates {(1.2, 0) (1.2, 1.2)};
        
        \end{axis}
        \end{tikzpicture}
        \[ \relu(x) = \left\{ \begin{array}{ll}
            x, & \mbox{$x \geq 0$}\\
            0, & \mbox{$x < 0$}\end{array} \right. 
        \]
        \caption{An illustration of the $\relu$ function. $p$ and $\Tilde{p}$ are on the two different branches of $\relu$. The slope between any two points on the function is always within $[0,1]$.}\label{fig:relu}
        \end{figure}


\paragraph{Shor's relaxation scheme}
We symbolize the computation components in the network and then constrain them with quadratic relations. The verification tasks are exactly encoded as QPs.
Unfortunately, QPs are generally $\NP$-hard to solve, because quadratic programs are quite expressive, and discrete conditions can be captured by them. For example, the MAXCUT problem can be easily expressed as a QP~\cite{maxcut}. To enable efficient solving, we relax the QP to the SDP that can be solved within polynomial time, using Shor's relaxation scheme~\cite{shor_SDP}. Shor's relaxation comes in two forms, which can be viewed as dual to each other~\cite{modern_co}.

The primal relaxation scheme is to relax each scalar variable to a multidimensional vector, and the dual form can be viewed as the Lagrangian relaxation of the original problem. In this work, we mainly use the primal form, so we provide some introduction here. The full detail of both forms of relaxation can be found in~\cref{sec:shor}.

Consider a general quadratic program:
\begin{align}\label{eq:quad-prog}
\begin{split}
    \min \;\;\;&f_0(x) = x^TA_0x+2b_0^Tx+c_0\\ 
    s.t.\;\;\;\;  & f_i(x) = x^TA_ix+2b_i^Tx+c_i \leq 0,\;\forall i\in[m]
\end{split}
\end{align}

We define a dyadic matrix $X(x) = \begin{pmatrix}
1\\
x
\end{pmatrix}\begin{pmatrix}
1\\
x
\end{pmatrix}^T$.

Then $x^T Ax + 2b^Tx+c = \begin{pmatrix}
1\\
x
\end{pmatrix}^T \begin{pmatrix}
&c \;\; &b^T\\
&b \;\; &A
\end{pmatrix}\begin{pmatrix}
1\\
x
\end{pmatrix} = \inprod{\begin{pmatrix}
&c \;\; &b^T\\
&b \;\; &A
\end{pmatrix},X(x)\Big}_F$.

$X(x)$ is the inner product of two vectors, so it is PSD and moreover a rank-1 matrix. If we drop the rank-1 requirement, we can get an SDP:
\begin{equation}\label{eq:prime-sdp}
    \min_{X}\{\inprod{\bar{A}_0,X}_F: \inprod{\bar{A}_i,X}_F\leq 0, i\in[m]; X\succeq 0; X_{11}=1 \},
\end{equation}
where
\[
\bar{A}_i=\begin{pmatrix}
c_i & \;\;\;\;b_i^T \\
b_i & \;\;\;\;A_i
\end{pmatrix}.
\]
The primal form of Shor's relaxation scheme can be viewed as the natural continuous relaxation for some combinatorial problems. We provide a discussion of this in~\cref{sec:shor}.

Given these components, we are ready to see how we can use the framework to verify the feed-forward DNN. We use the two-layer network as an example, and it is straightforward to extend to multi-layer networks within the framework. Let us consider a two-layer network: $f:\R^m\rightarrow \R$, with one hidden layer of dimension $n$:
\begin{equation}\label{eq:2-network}
   f(x) = v\act(Wx+b)+c, 
\end{equation}

where $W\in \R^{n\times m}$, $b\in \R^{n\times 1}$, $v\in \R^{1\times n}$ and $c\in \R$. Let $w_i$ be the $i$-th row vector of $W$.

Given Shor's relaxation scheme and SDP solvers, verifying neural network properties amounts to formulating the tasks as QPs.

\section{Methodology}\label{sec:method}
In this section, we elaborate on the symbolic reasoning framework for some common feed-forward network verification tasks with $\ell_p$-perturbations. These are considered as the standard DNN verification problems. 
We first present an overview of the symbolic reasoning framework, and then show how we can concretely apply this framework to verify DNNs in different scenarios.

\subsection{Overview}\label{sec:overview}
Similar to the classical symbolic execution~\cite{symbex}, the framework ``executes'' the DNN and generates an optimization program that precisely characterizes the problem of interest\footnote{We categorize the satisfiability program as an optimization program in this paper. For example, it is easy to transform a SAT problem into a MaxSAT problem.}. For this step, the challenge is how we can execute the DNN symbolically and also encodes the verification problem that we aim to address. We utilize the expressiveness of quadratic relations, and the resulting program is a QP rather than a satisfiability program in the classical symbolic execution.

In the classical symbolic execution, once the satisfiability program is generated, one can use off-the-shelf solvers such as the SMT solver to solve the program~\cite{smt}. Our framework contains an extra relaxation step to achieve \emph{efficiency}. Many DNN verification tasks are known $\NP$ or $\coNP$-hard~\cite{reluplex,IUA,geolip}. Because the QPs exactly encode the problems of interest, they are also hard to solve. However, we want to address the verification problem efficiently, i.e., in polynomial time, so we need to relax the QP. We use Shor's relaxation scheme, to transform the hard-to-solve QP to the convex SDP. SDPs can be solved efficiently both theoretically and in practice.

\paragraph{Neural-network execution}A feed-forward network is a composition of affine transformations and non-linear activations. To reason it symbolically, we associate symbols with the input and the output of any component function, and then relate those symbols with quadratic relations. See~\cref{fig:network} for an example.

Therefore, we only need to symbolically execute the affine transformation and the activation function. For affine transformations, it is straightforward because affine relations are linear and so quadratic. The interesting part is how to execute the activation function. Suppose $z=\relu(y)$, then we can have the following different interpretations of $\relu$'s execution.

\tikzset{%
  every neuron/.style={
    circle,
    draw,thick,
    minimum size=0.5cm
  },
  neuron missing/.style={
    draw=none, 
    scale=4,
    text height=0.333cm,
    execute at begin node=\color{white}
  },
}

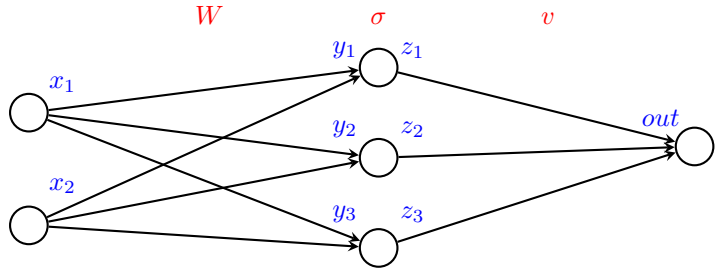
\begin{figure}
    \footnotesize
    \centering
    \begin{tikzpicture}[x=1.5cm, y=1.5cm, >=stealth]

        \foreach \m/\l [count=\y] in {1,2}
          \node [every neuron/.try, neuron \m/.try] (input-\m) at (-0.1,2.6-\y) {};
        
        \foreach \m [count=\y] in {1,2,3}
          \node [every neuron/.try, neuron \m/.try ] (hidden1-\m) at (3,2.8-\y*0.8) {};
        
        \foreach \m [count=\y] in {1}
          \node [every neuron/.try, neuron \m/.try ] (output-\m) at (5.8,2.3-\y) {};
        
        

        \foreach \i in {1,...,2}
          \foreach \j in {1,...,3}
            \draw [->,thick] (input-\i) -- (hidden1-\j);

        \foreach \j in {1,...,3}
            \draw [->,thick] (hidden1-\j) -- (output-1);

        

        
        
        
        

        \node [align=center, above, color=blue] at (0.2, 1.7) {$x_1$};

        \node [align=center, above, color=blue] at (0.2, 0.8) {$x_2$};

        \node [align=center, above, color=blue] at (2.7, 1.3) {$y_2$};

        \node [align=center, above, color=blue] at (2.7, 2.0) {$y_1$};

        \node [align=center, above, color=blue] at (2.7, 0.55) {$y_3$};

        \node [align=center, above, color=blue] at (3.3, 1.3) {$z_2$};

        \node [align=center, above, color=blue] at (3.3, 2.0) {$z_1$};

        \node [align=center, above, color=blue] at (3.3, 0.55) {$z_3$};

        \node [align=center, above, color=blue] at (5.5, 1.4) {$out$};

        \node [align=center, above, color=red] at (1.5, 2.3) {$W$};
        \node [align=center, above, color=red] at (4.5, 2.3) {$v$};
        \node [align=center, above, color=red] at (3, 2.3) {$\act$};
        
        
        
        \end{tikzpicture}
        \caption{A two-layer network example: 
$f(x) = v\act(Wx+b)+c$. We use the symbol $x$ to denote the input of the network, $y$ for the input of the hidden layer, $z$ for the output of the hidden layer, and $out$ as the out symbol. The subscript denotes the component within each symbol.}\label{fig:network}
     
\end{figure}

(\textbf{i}). The first interpretation is based on the observation that $\relu$'s slope is always within $[0,1]$, i.e., for any two inputs $y$ and $\Tilde{y}$ with $z = \relu(y)$ and $\Tilde{z} = \relu(\Tilde{y})$:
\[0 \leq \frac{z-\Tilde{z}}{y-\Tilde{y}}\leq 1.\]
This can be encoded with a quadratic relation: $((z-\Tilde{z})-(y-\Tilde{y}))(z-\Tilde{z})\leq 0$. This ReLU interpretation is used in~\cite{lipsdp}, and was named as the \emph{slope-restricted} activation. See~\cref{fig:relu} for an example. 

(\textbf{ii}). The second interpretation is from~\cite{sdp_rob_local}. $\relu$'s computation can be captured by the following three quadratic inequalities: $y\leq z$, $z\geq 0$ and $(z-y)z \leq 0$. One can easily verify that this encodes the computation of $\relu$ exactly. 

\begin{remark} 
It is to see that the slope-restricted interpretation is not unique to $\relu$, i.e., any function whose slope/derivative is between $0$ and $1$ has the same interpretation as $\relu$.
\end{remark}

\begin{remark}\label{rm:unique} 
The precise quadratic interpretation of $\relu$ is not unique. For example, we can introduce a new symbol $s$ to denote which branch of $\relu$ the execution is on. To express $\relu$'s computation, we can use $s(s-1) = 0$, $(s-1/2)y\geq 0$ and $z=sy$. This interpretation is similar to the one in~\cite{chen2020semialgebraic}. From this interpretation, one can view the slope-restricted interpretation as a \emph{stateless} interpretation of $\relu$, which does not consider the execution status of $y$.
\end{remark}
We also provide quadratic encodings for some other activation functions in~\cref{sec:other-act}. In particular, for those non-algebraic activation functions, we can use results from approximation theory to handle them.

\paragraph{Input perturbation}
Commonly considered attacks on the input are $\ell_2$ and $\ell_\infty$-attacks. Here we present how to encode them using quadratic relations. For other $\ell_p$-perturbations, we show their encodings in~\cref{sec:ell-p}.
Suppose that we have two inputs $x, \Tilde{x}\in \R^m$, if their $\ell_2$-distance is within $\epsilon$, then we can write it as:
\[
\sum_{i=1}^m (x_i-\Tilde{x}_i)^2 \leq \epsilon^2.
\]
If the $\ell_\infty$-distance between $x$ and $\Tilde{x}$ is within $\epsilon$, then $\max_i|x_i-\Tilde{x}_i|\leq \epsilon$. This can be encoded as $(x_i-\Tilde{x}_i)^2\leq \epsilon^2$ for all $i\in [m]$.


\subsection{Data-dependent Analysis}\label{sec:dept}
For data-dependent analysis, or local analysis, we fix an input $a\in \R^m$. We want to quantify the change of $f(a)$ if an $\ell_p$-perturbation with radius $\epsilon$ is added to the input. As a result, the quadratic program for the problem is:
\begin{subequations}
\begin{align}
    \max &\;vz+c \label{eq:obj}\\ 
    s.t.\;\;\;\;  & z_i(z_i-y_i)\leq 0, \;z_i\geq y_i, \;z_i\geq 0, \;\forall i\in [n]  \;\;(\relu \text{ computation}) \label{eq:relu}\\
    &y_i = w_i x+b_i,\;\forall i\in [n] \label{eq:first-layer}\\
    &\norm{x-a}_p\leq \epsilon \label{eq:pert}.
\end{align}
\end{subequations}
The semantics of the symbols are the same as in~\cref{fig:network}. \Cref{eq:pert} denotes that $x$ is within the $\ell_p$-ball centered at $a$ with radius $\epsilon$. \Cref{eq:first-layer} denotes the affine transformation of the first layer, which constrains $x$ and $y$.  \Cref{eq:relu} denotes the element-wise computation of the $\relu$ hidden layer, which constrains $y$ and $z$. \Cref{eq:obj} is the objective that quantifies the output change. Because each equation exactly expresses the corresponding computation, this QP precisely encodes the verification problem of interest.

Now we demonstrate how to relax the quadratic program for $\ell_2$-perturbations using Shor's relaxation. The SDP for $\ell_\infty$-perturbations can be found in~\cite{sdp_rob_local}. The $\ell_2$-ball centered at $a$ can be expressed as
$\sum_{i=1}^m (x_i-a_i)^2\leq \epsilon^2$.
To relax the quadratic program to an SDP in the primal form, we first define a PSD variable $V\succeq 0$ with the decomposition as:
\[
V = \begin{pmatrix}
1 \;\;& z \;\;& x \\
z^T \;\;& Z \;\;& Y \\
x^T \;\;& Y^T \;\;& X \\
\end{pmatrix}
\]
Because $V$ is PSD, then $\exists M\in \R^{l\times (1+n+m)}$ for some $l\in\Z_+$ such that $V = M^TM$. Let us write $M = \begin{pmatrix}
u\;\; \bar{z}\;\; \bar{x}
\end{pmatrix}$ with $u^Tu = 1$. We can think of $u$, $\bar{z}$ and $\bar{x}$ as the multi-dimensional relaxations of $1$, $z$ and $x$. Representing the variables in $V$ with $M$'s components, we have $z = u^T\bar{z}$, $Z = \bar{z}^T\bar{z}$, $x = u^T\bar{x}$, $X= \bar{x}^T\bar{x}$, and $Y = \bar{z}^T\bar{x}$.

Because $y_i = w_ix+b_i$, we can substitute $y_i$ with $w_ix+b_i$. Now we are ready to present the SDP:
\begin{align}\label{eq:l2-cert-sdp}
\begin{split}
    \max &\;v z+c \\ 
    s.t.\;\;\;\;  & Z_{ii} - b_iz_i - w_iY_i^T \leq 0, \;z_i\geq w_ix+b_i, \;z_i\geq 0, \;\forall i\in [n] \\
    & \trace{(X)} - 2 x^T a + a^Ta - \epsilon^2\leq 0,
\end{split}
\end{align}
where $Y_i$ is the $i$-th row of $Y$.

For multi-layer networks, to construct the QP that expresses the verification problem, we only need to introduce more symbols to denote the input and output of the layers, and then constrain them using the quadratic relations as in the two-layer DNN case. It is then straightforward to relax the QP to SDP using Shor's relaxation. 

\subsection{Data-independent Analysis}\label{sec:indept}
The data-independent analysis aims to quantify the change of a DNN, given a data-independent perturbation. In other words, we want to upper bound the change of the DNN output if a perturbation can be applied at any input point. This property is intrinsic to the neural network as a function, and not dependent on the data; and the problem is also known as the Lipschitz constant estimation of $f$. From mathematical analysis, the Lipschitz constant is the maximum operator norm over all possible gradients. 
\cite{geolip} defined the formal global Lipschitz constant (FGL) as the maximum operator norm of all possible gradients, in which all the activation functions are considered independent. In reality, the activation functions are correlated and some of the activation patterns are infeasible. As a result, this value is a formal quantity and upper bounds the Lipschitz constant. \cite{geolip} indicated that most of the works on Lipschitz estimation and regularization, such as~\cite{alg-lip,lipsdp,gloro,certified_def} study the FGL instead of the Lipschitz constant. The study of FGL, especially the SDP estimation of this quantity~\cite{lipsdp}, inspires~\cite{alg-lip} to design $1$-Lipschitz network structures.

Using SDP to estimate the DNN Lipschitz constant was initiated by \cite{lipsdp}\footnote{\cite{certified_def} also devised an SDP to estimate the formal Lipschitz constant but the approach only works for two-layer networks.}, which only works for $\ell_2$-perturbations. \cite{lipopt} claimed that~\cite{lipsdp} could not transfer to the $\ell_\infty$ case. \cite{geolip} gave a compositional interpretation of~\cite{lipsdp} and generalized~\cite{lipsdp} to the $\ell_\infty$ case.
In fact, the SDPs for FGL estimation are just Shor's relaxation of QPs with the stateless $\relu$-interpretation. 

\paragraph{Formal Lipschitzness estimation}
In FGL estimation, all activation are considered independent, therefore, one does not need to consider the actual execution state of the $\relu$ function. \cite{geolip} used $\Delta x$ to denote the difference between two inputs $x$ and $\Tilde{x}$, and similarly $\Delta y$ for $y$ and $\Tilde{y}$, $\Delta z$ for $z$ and $\Tilde{z}$. The following QP can encode the FGL-estimation problem:
\begin{align}\label{eq:formal}
\begin{split}
  \max &\;v\Delta z \\ 
    s.t.\;\;\;\;  & (\Delta z_i - \Delta y_i)\Delta z_i\leq 0, \;\forall i\in [n] \\
    & \Delta y_i = w_i \Delta x,\;\forall i\in [n]\\
    &\norm{\Delta x}_p\leq 1  .
\end{split}  
\end{align}
An interpretation of this program is to quantify how a data-independent perturbation $\Delta x$ propagates from the input layer to the output layer. The propagation is subject to the $\ell_p$-norm and DNN computation constraints.
\cite{geolip} provided more details on the relaxed SDPs of~\cref{eq:formal}, the relations to the mixed-norm problem and the Grothendieck inequalities~\cite{cut-norm,GroIneq,nest}. In particular, the SDPs in \cite{lipsdp} and~\cite{certified_def} are just Shor's relaxations of~\cref{eq:formal} within their application scope in the primal and dual forms.

\section{Power of Symbolic Reasoning}\label{sec:power}
In this section, we examine a few tasks beyond the feed-forward network verification. These tasks appear challenging for non-symbolic domains, but can be naturally solved with our framework. We can assign any computational component in the network with a symbol on demand without defining the abstract interpreter a priori. This flexibility contributes to the power of symbolic reasoning.

The major distinguished components of our framework are the symbolic domain and the quadratic relation. The symbolic domain has very flexible semantics. We can assign symbols to any computation component in the network, and define their semantics as needed. In the meantime, the quadratic relation is quite expressive, allowing us to exactly encode the verification problem of interest. We can then use algebraic manipulations and other mathematical tools to precisely analyze those problems. Moreover, using quadratic relations to exactly express the verification problem enables us to ask important theoretical questions about the verification task, which we will discuss in~\cref{sec:theory}. Now we demonstrate the practical benefits with the following examples.

\subsection{Metric Learning}
Metric learning models learn a metric that captures the semantic features. They map inputs into a low-dimensional space on which distance measures the similarity. This model is widely used in face recognition, information retrieval and phishing detection~\cite{phishNet,facenet,Wu_2017_ICCV}. The model structure is essentially the same as the feed-forward except for the final classification layer. Instead of outputting the class with the highest logit score, the model maps the output from the representation layer onto a unit sphere, and outputs the closest anchor's class. 

More formally, let $z$ be the output of the representation layer, and $\bar{z}$ be the normalized $z$, i.e., $\bar{z} = \frac{z}{\norm{z}_2}$. $O = \{o_i\}$ is a set of anchors on the unit sphere (so $\norm{o_i}_2=1$). To predict $x$, the model examines the $\ell_2$-distance between $\bar{z}$ and all the anchors, and outputs the class of the closest anchor.  

\cite{adv-metric} found that metric learning models are also susceptible to adversarial attacks. It appears difficult to verify the metric learning model using non-symbolic domains. Reinterpreting the normalization operation and the $\ell_2$-distance on a unit sphere can be particularly challenging for those domains. However, we demonstrate that with symbolic reasoning, verifying metric learning models is no different from the standard feed-forward model.

Let $o$ be the closest anchor, and $\Tilde{o}$ be another arbitrary anchor. Therefore, we want to minimize $(\bar{z}-\Tilde{o})^2-(\bar{z}-o)^2$ to see if this expression can be negative. With some arithmetical operations:
\[(\bar{z}-\Tilde{o})^2-(\bar{z}-o)^2 = \bar{z}^2+\Tilde{o}^2-\bar{z}^2-o^2+2(o-\Tilde{o})\bar{z}.\]

Because $o^2, \Tilde{o}^2, \bar{z}^2=1$, this is equivalent to minimizing $(o-\Tilde{o})\bar{z}$. Geometrically, this means whether the angle between $\bar{z}$ and $(o-\Tilde{o})$ is greater than $\frac{\pi}{2}$. Therefore, this is equivalent to whether $z$ can be perturbed to the other side of the half-space that is normal to $(o-\Tilde{o})$. 

Equivalently, we only need to minimize $(o-\Tilde{o})z$ to see whether this can be negative. Now assuming the metric learning model has only two layers as in~\cref{eq:2-network}, the quadratic program for metric learning is then:
\begin{align*}
    \min &\;(o-\Tilde{o})z\\ 
    s.t.\;\;\;\;  & z_i(z_i-y_i)\leq 0, \;z_i\geq y_i, \;z_i\geq 0, \;\forall i\in [n] \\
    &y_i = w_i x+b_{1i},\;\forall i\in [n] \\
    &\norm{x-a}_p\leq \epsilon .
\end{align*}

With symbolic reasoning, we can conclude that verifying metric learning models is essentially the same as the standard classification model. We can use a vector $v$ to denote $\Tilde{o}-o$, then the program is the same as in the standard feed-forward model.

\subsection{Deep Equilibrium Models}
Symbolic abstraction is a fundamental reasoning scheme and is used beyond program verification. It can be particularly elegant when reasoning the limiting behavior of recursions. We provide a probability theory example to show this reasoning in~\cref{sec:rec-prob}. We assign the symbol with semantics in the limiting state directly without handling fixed points. Now we demonstrate this reasoning in the verification setting.

We show that we can easily reason about the deep equilibrium (DEQ) model~\cite{deq_model}. DEQs use a single implicit layer to simulate a network with infinite depth. It is based on the observation that very deep models converge towards some fixed point, and the implicit layer solves this fixed point directly. It has been shown that DEQs have competitive performance compared to explicit deep models while consuming much less memory~\cite{multi-deq}.

The DEQ model $f:\R^m\rightarrow \R$ can be described with the following formula:
\[f(x) = vz+b_2, \;z = \act(Wz+Ux+b_1).\]
If we compare the DEQ model with the feed-forward model (see~\cref{eq:2-network}), the difference is that $z$ is also fed back to the input of the hidden layer, i.e., $z$ is a fixed point of the hidden layer equation. The model then uses the affine transformation of this fixed point to make predictions, i.e., $\class(f,x)=\argmax_{i\in[l]}Vz+b$, where $V$ is the weight matrix from the fixed point $z$ to the logit layer, and $b$ is the bias term.

If we assign a non-symbolic domain to capture the semantics of $x$, reasoning $z$ requires finding the fixed point within this domain, which can be very loose. Instead, we can assign symbols to denote the equilibrium behavior directly. We use the FGL estimation (see~\cref{sec:indept}) as an example. 

To derive the QP like~\cref{eq:formal}, the only difference between the DEQ model and the feed-forward DNN is that $\Delta y_i = w_i\Delta z + u_i\Delta z$, where $u_i$ is the $i$-th row of $U$. In other words, an input to the hidden layer ($\Delta y_i$) comes from both the input layer ($w_i\Delta x$) and the output of the hidden layer ($u_i\Delta z$). The rest remains the same. As a result, the QP for the FGL estimation of DEQ is:
\begin{align*}
\begin{split}
  \max &\;v\Delta z \\ 
    s.t.\;\;\;\;  & (\Delta z_i - \Delta y_i)\Delta z_i\leq 0, \;\forall i\in [n] \\
    & \Delta y_i = w_i \Delta x+u_i\Delta z, \;\forall i\in [n]\\
    &\norm{\Delta x}_p\leq 1.
\end{split}  
\end{align*}

\section{Evaluation}\label{sec:eva}
We want to understand the quality of the framework proposed in the work. Since this is a framework, it can be applied to different tasks. We choose $\ell_2$-data-dependent analysis for evaluation because there are no tools amenable to our framework but there are other benchmark tools that we can compare with. This can help us evaluate the empirical strength and weakness of our framework. 

Specifically, we want to certify the robustness of data when $\ell_2$-perturbations of a specific radius are allowed. The SDP for this task is described in~\cref{eq:l2-cert-sdp}. On this certification task, we aim to empirically answer the following research questions:
\begin{tcolorbox}
\begin{enumerate}[start=1,label={\bfseries RQ\arabic*:}]
\item How precise is the framework to certify inputs subject to $\ell_2$ adversarial attacks?
\item How efficient is the framework to certify data points?
\end{enumerate}
\end{tcolorbox}
Notice that RQ1 is usually problem specific, independent of the computing environment and implementation if unlimited computing resources are assumed, but RQ2 heavily relies on implementation and computing environment. On average, the framework is $60\%$ better than the benchmark tool, and the result is summarized in~\cref{tab:result}. However, because the framework relies on off-the-shelf SDP solvers, in our implementation and experiment, the framework is considerably slower than the benchmark, as summarized in~\cref{tab:time-result}. However, our implementation is still able to handle practical-sized networks, compared to exponential-time verifier, for example, Reluplex~\cite{reluplex}.

\paragraph{Server specification}All the experiments are run on a workstation with forty-eight Intel\textsuperscript{\textregistered} Xeon\textsuperscript{\textregistered} Silver 4214 CPUs running at 2.20GHz, and 258 GB of memory, and eight Nvidia GeForce RTX 2080 Ti GPUs. Each GPU has 4352 CUDA cores and 11 GB of GDDR6 memory.

\paragraph{Tools}Our framework transforms the verification tasks into SDPs, and to finish the analysis, we need to rely on off-the-shelf SDP solvers. We implement our framework for $\ell_2$-local analysis using the MATLAB CVX and MOSEK solver~\cite{MATLAB,cvx,mosek}, and name the tool \emph{ShenBao}, which resembles ``symbol'', to emphasize the symbolic reasoning essence.

We use BCP~\cite{bcp} as the benchmark to certify the robustness of inputs. BCP uses both the $\ell_2$-Lipschitz constant and the interval domain to refine the abstract interpretation of the input execution induced from $\ell_2$-balls.

Additionally, we also use the PGD attack~\cite{madry2018towards} to evaluate the model. PGD adversarial examples are considered standard $\ell_2$ and $\ell_\infty$-attacks, and are also used in adversarial training. The result from PGD attacks can also serve as a sanity check for certifiable correctness because certifiable correctness by definition is no higher than adversarial accuracy for any attacks.

\paragraph{Attacks} The \emph{strength} of attack is measured by the attack radius. For robust and adversarial training, we train the model with one strength and test the model using a weaker, the same, and a stronger attacks to measure the performance. We use $\epsilon$ to denote the attack strength during training. 

\paragraph{Neural network models}We use two models for evaluation. One is a \emph{small} DNN, consisting of a single hidden layer with $64$ $\relu$ nodes; and the other one is a \emph{medium} DNN with two hidden layers: the first layer has $128$ nodes and the second one has $64$ nodes. 

We train our network on the MNIST dataset~\cite{mnist}. The models are trained under three different modes. The first one is the standard natural training. The second one is the BCP training as in~\cite{bcp}. The third one is PGD training: the model is fed with PGD adversarial examples of certain attack strengths during training.

For all small models, we train the model for $60$ epochs; and for all medium models, we train the model for $100$ epochs.

\paragraph{Measurement}We fix $200$ test data points that are inaccessible to the model during training. In the test phase, we measure the following quantities:
\begin{enumerate}
    \item \emph{Accuracy} denotes how many inputs can be correctly classified when no attacks are applied;
    \item \emph{PGD} measures how many points can still be correctly classified when PGD attacks are applied.
    \item \emph{ShenBao} denotes how many inputs can be certifiably classified correctly by ShenBao when $\ell_2$-attacks are allowed.
    \item \emph{BCP} measures how many points can be certifiably classified correctly by BCP when $\ell_2$-attacks are allowed.
\end{enumerate}

\paragraph{Precision of certification} The result of certification precision is summarized in~\cref{tab:result}. 

\begin{table}
\begin{center}
\begin{tabular}{p{3cm}p{2cm}p{1.8cm}p{1.5cm}p{1.8cm}p{1.5cm}}
\toprule
\multicolumn{6}{c}{\centering Number of Correctly Classified Inputs} \\
\midrule
\centering Model & \hfil Accuracy &\hfil Strength & \hfil PGD & \hfil ShenBao & \hfil BCP \\
\midrule
 \centering Small & \hfil  & \hfil 0.3  & \hfil 181& \hfil 126 & \hfil 58 \\
\centering Natural Training & \hfil 200 & \hfil 0.5 & \hfil 138& \hfil 70 & \hfil 5 \\
\centering$\epsilon=0$ & \hfil  & \hfil 0.7  & \hfil 90& \hfil 20 & \hfil 0 \\
\midrule
\centering Small &\hfil   & \hfil 0.3 & \hfil 192& \hfil 186  & \hfil 188  \\
\centering BCP Training & \hfil 198 & \hfil 0.5 & \hfil 190& \hfil 177 & \hfil 176 \\
\centering $\epsilon=0.5$ & \hfil  & \hfil 0.7 & \hfil 182& \hfil 156 & \hfil 158 \\
\midrule
\centering Small & \hfil  & \hfil 0.3 & \hfil 196& \hfil 193 & \hfil 159  \\
\centering PGD Training & \hfil 198 & \hfil 0.5 & \hfil 191& \hfil 178 & \hfil 72 \\
\centering $\epsilon=0.5$ & \hfil & \hfil 0.7 & \hfil 182& \hfil 161 & \hfil 12 \\
\midrule
\centering Medium & \hfil  & \hfil 0.7& \hfil 182& \hfil  151& \hfil 158  \\
\centering BCP Training & \hfil 193 & \hfil 1.0 & \hfil 166& \hfil 126 & \hfil 128 \\
\centering $\epsilon=1.0$ & \hfil & \hfil 1.3 & \hfil 146& \hfil 100 & \hfil 105 \\
\midrule
\centering Medium &\hfil  & \hfil 0.7& \hfil 195& \hfil 177 & \hfil 0  \\
\centering PGD Training & \hfil 199  & \hfil 1.0& \hfil 179& \hfil 142 & \hfil 0 \\
\centering $\epsilon=1.0$ & \hfil & \hfil 1.3& \hfil 167& \hfil 90 & \hfil 0 \\
\bottomrule
\end{tabular}
\newline\newline 
\caption{The result of experiments to certify the robustness of $200$ test inputs. The column PGD denotes how many points remains accurate with PGD attack. ShenBao denotes how many points are certified accurate by ShenBao. BCP denotes how many points are certified accurate by BCP. We can see that ShenBao performs consistently better than BCP except for a few cases when the network is trained with the BCP-induced bound. It is noteworthy that on the medium PGD-trained model, it is already quite certifiably robust from the measurement of ShenBao while BCP cannot certify the robustness at all. }\label{tab:result}
\end{center}
\end{table}

Additionally, we also profile the time used by ShenBao and BCP to certify the inputs.

\paragraph{Speed of certification}
We measure the certification speed of ShenBao and BCP. We ran our experiments on the MNIST dataset, which has $10$ classes. For ShenBao to certify an input, we need to solve $(10-1=9)$ SDP programs. These $9$ SDP programs are independent, so can be solved in parallel. We record the average running time to solve one SDP for both the small and the medium models.

On the other hand, BCP is implemented to run on GPUs very efficiently. We record the total time to certify all $200$ data. The certification time is summarized in~\cref{tab:time-result}.

\begin{table}
\begin{center}
\begin{tabular}{p{4cm}p{3cm}p{3cm}}
\toprule
\multicolumn{3}{c}{\centering Running Time (in seconds) of ShenBao and BCP} \\
\midrule
\centering Model & \hfil ShenBao &\hfil BCP \\
\midrule
\centering Small & \hfil 102.4 &\hfil  2.1\\
\midrule
\centering Medium & \hfil 426.2 &\hfil  1.8\\
\bottomrule
\end{tabular}
\newline\newline 
\caption{The running time of certification. On average, ShenBao needs $102.4$ seconds to solve the SDP from a small model, and $426.2$ seconds to solve the SDP from a medium model; while BCP only needs about $2$ seconds to certify all $200$ inputs.}\label{tab:time-result}
\end{center}
\end{table}

\section{Discussion}\label{sec:dis}
We provide some discussion on the empirical result and the theoretical aspects of the framework.
\subsection{Empirical Discussion}
\paragraph{RQ1}
From~\cref{tab:result}, we can conclude that ShenBao performs consistently better than BCP, except for a few cases when the model is BCP-trained. BCP-training will regularize the model to have a tighter BCP bound, and thus benefits the BCP certification. Even on these models, ShenBao still achieves comparable performance. This demonstrates the good precision of our framework.

One interesting observation is that networks not trained for certifiably robust purposes are also certifiably robust to some extent from the measurement of ShenBao. We would not observe this with only the measurement from BCP. This is particularly true for the PGD-trained medium model. The certifiable robustness of the model is about $88\%$ with an attack radius of $0.7$, while BCP is unable to certify the robustness at all. One surprising observation is that PGD training was not considered certifiably robust because the PGD attacks are weaker than the certifiably robust bound. However, with ShenBao's measurement, PGD-trained models can achieve similar or even better certifiable robustness compared to BCP-trained models.

\paragraph{RQ2}
The bottleneck for using our framework is solving the SDPs generated from the verification tasks. In our implementation, we use the MOSEK solver, a generic SDP solver. BCP is implemented to use GPUs, and can run very fast, while the MOSEK solver that we use for the evaluation can only run on CPUs.

Because all the SDPs are independent, one can solve all the verification SDPs in parallel when more CPUs are available. There are a few other works on improving the SDP solving for DNN certification. \cite{efficient_sdp} devised a first-order algorithm to solve the SDP that is implementable on GPUs, and successfully scaled the SDP verification to CIFAR-size networks. \cite{chordal_sparsity,explore_chordal} explored the chordal sparsity manifested in the DNN-induced SDPs. The chordal sparsity enables decomposing the large PSD constraint into a few smaller ones, so solving the SDP is much faster. However, solving the SDPs induced by the verification tasks is beyond the scope of this work. Our work demonstrates the power of SDPs in verifying DNNs, and can motivate more future work on specified SDP-solving algorithms for DNN verification.

\paragraph{Practical implication}
Unlike theoretical exponential-time verification tools, e.g., Reluplex~\cite{reluplex}, our framework is theoretically polynomial-time and can handle reasonable real-world size networks like the MNIST network. Moreover, it is powerful enough to reason about various network models and properties beyond the linear-constrained properties on feed-forward networks. 

Our framework complements the current verification methodology. For example, with symbolic reasoning, we can conclude that certifying metric learning models is no different from the feed-forward model. One can then choose techniques from the feed-forward models to verify the metric learning models.

On the other hand, for verification tasks or models without other techniques available, one can try to use quadratic relations to represent the verification task and then relax the program to an SDP. On small models, our framework can provide useful verification measurements. When more efficient and scalable methods are available, our method can also serve as a benchmark that provides information about the more efficient methods. 

\subsection{Theoretical Discussion}\label{sec:theory}
\paragraph{Other relaxations}We broadly categorize any reasoning method, which results in algebraic expressions, to the symbolic reasoning paradigm. The decision or numerical results are usually not immediately available from these expressions so they need further analysis. Therefore, one can view any DNN verification works that require solvers to post-process the algebraic expressions as examples of symbolic reasoning. In this work, we used quadratic encodings and Shor's relaxation to achieve efficiency. One can also consider other symbols and relational representations to express and relax the verification problem.~\cite{reluplex} used Reluplex to exactly verify neural networks.~\cite{linear-prog} used linear programming (LP) to relax the certification problem. Because many verification problems are intrinsically non-linear, one has to relax the problem to enable efficient verification. 

\cite{int-lp} demonstrated that there is an intrinsic precision gap between polynomial-sized LPs and SDPs on some natural optimization problems. This implies that SDP is strictly more powerful than LP in solving intractable problems within polynomial time. It is interesting to understand whether a similar gap exists for DNN verification problems. However, without the QP representations, we would not be able to ask this question.

\paragraph{Quadratic encoding}
Quadratic encoding is fairly expressive. For example, $x(x-1)=0$ can express $x=0$ or $x=1$. This enables the quadratic program to encode many intractable combinatorial problems. See~\cite{modern_co} for more information on the expressiveness of quadratic relations. In the meantime, the quadratic encoding of a problem is not necessarily unique. For example, we presented two exact quadratic encodings for $\relu$ (see~\cref{rm:unique}). This is important when we consider its SDP relaxation because equivalent quadratic programs can have different relaxations, which result in different solutions~\cite{sdp-relax-operator}. Exploring which encoding gives better relaxation is an interesting question but it is beyond the scope of this work. 

\paragraph{SDP relaxation}
Using SDP relaxation to solve intractable problems is pioneered by the seminal Goemans-Williamson algorithm for the MAXCUT problem~\cite{maxcut}. It is an approximation algorithm with a tight bound, and is the optimal polynomial-time algorithm assuming the unique games conjecture~\cite{UGC,ugc_cut}.

As such, an important theoretical problem for the SDP framework is how precise the relaxation is. \cite{sdp_quality} studied cases when the SDP relaxation is precise, i.e., has no precision loss for certifying the robustness of inputs.

\cite{geolip} proved that on two-layer networks, the SDP relaxations have approximation guarantees of $K_G$ (known as the Grothendieck constant) for $\ell_\infty$-FGL and $\sqrt{\pi/2}$ for $\ell_2$-FGL on two-layer networks.
These guarantees rely on the results from the mixed-norm problem. The $p\rightarrow q$ mixed-norm of a matrix $A$ is defined as $\max_{\norm{x}_p = 1}\norm{Ax}_q$.

\cite{cut-norm} proved that the SDP for the $\infty\rightarrow 1$ mixed-norm problem induces an $K_G$-approximation guarantee. The $\infty\rightarrow 2$ mixed-norm problem has an $\sqrt{\frac{\pi}{2}}$ approximation guarantee from its SDP. $\frac{\pi}{2}$ comes from Grothendieck's original paper~\cite{GroIneq} and was rediscovered by~\cite{nest} later. \cite{geolip} built reductions from the FGL estimation to the mixed-norm problem, so the FGL estimation admits the same approximation ratios as the mixed-norm problem.
Moreover, \cite{opt_inf2} showed that assuming $\p\neq \NP$, the $\sqrt{\frac{\pi}{2}}$-approximation guarantee is optimal for the $\infty\rightarrow 2$ mixed-norm problem; and \cite{UGC_inf1_opt} showed that assuming the unique games conjecture, the $K_G$-approximation guarantee is optimal for the $\infty\rightarrow 1$ mixed-norm problem. 

However, there are no approximation guarantees for multi-layer DNN FGL-estimation SDPs, and the authors of~\cite{geolip} posed this problem as an open question.

\paragraph{SDP representation of verification problems}
One perspective of this framework is that the SDP provides a new representation of the verification task. SDP is intrinsically connected to many subjects, so the new representation also establishes relations between the verification problem and other mathematical tools.
As an analogy, in spectral graph theory~\cite{spectra}, one can study the matrix representation of a graph and derive graph properties from these matrices. 

Here we prove an analytical result about the SDPs for FGL estimations on two-layer networks. This exemplifies how the new representation brings new techniques to study the verification problem.

For a two-layer network, the primal SDPs for FGL estimation are of the following form~\cite{geolip}:
\begin{align}\label{eq:fgl-sdp}
\begin{split}
   \max\;\; &\inprod{M,X}_F\\
    s.t.\; & X \succeq 0, X_{ii}=1, i\in [n],
\end{split}
\end{align}
where $M\in \R^{n\times n}$ is a symmetric matrix for some $n\in\Z_+$.

Let $h\in \R^{n}$ be a vector whose entries sum to $0$, i.e., $\sum_{i=1}^{n} h_i = 0$. If we use $O$ to denote the optimal value of~\cref{eq:fgl-sdp}, we have the following result:

\begin{theorem}\label{thm:weighted-avg}
$O = \min_{h}n\eig_{\max}(M+\ACTL(h))$.
\end{theorem}
That is to say, to estimate a precise FGL of the network, one only needs to check the largest eigenvalue of the $M+\ACTL(h)$. This result is motivated by~\cite{lap_eigen}. \cite{certified_def} used a similar technique to train robust networks subject to the SDP constraint. Our result can be viewed as an improvement of theirs. See~\cref{sec:proof} for more discussion.

As we can see from this result, the SDPs contain very rich information about the verification task, especially from the spectrum of the matrices. Therefore, the SDP programs provide new representations of the verification problems, and exploiting the SDP structures brings more understanding of these problems.

Meanwhile, the framework is expressive and can encode many network properties. The resulting SDPs can provide estimations of these properties. It is appealing to train networks with the constraints from the SDPs because this can regularize the network to achieve desired properties. For this purpose, it is promising to analyze the SDPs and discover surrogates amenable to the first-order method.  

\paragraph{Symbolic framework}
One way of viewing our work is to distill a framework from a few existing examples and generalize this framework to other verification tasks. We argue that this distillation is important and necessary. For example, in~\cref{sec:indept}, we mentioned that~\cite{geolip} showed that the SDPs in~\cite{lipsdp} and~\cite{certified_def} are just Shor's relaxations of~\cref{eq:formal} in the primal and dual forms. \cite{lipsdp} works for multi-layer network $\ell_2$-FGL estimation, and~\cite{certified_def} is designed for two-layer networks when $p=\infty$. However, this duality is not clear without demonstrating the framework. For example, \cite{lipopt} argued that~\cite{lipsdp} does not work when $p=\infty$, and~\cite{sdp_rob_local} believed that~\cite{certified_def} only works for two-layer networks. Nevertheless, the SDPs are general enough, and this is straightforward from our framework. Similarly, we implemented a tool for $\ell_2$-robustness certification, which is natural from our framework. However, even though it has been a while since the SDP for $\ell_\infty$-robustness certification was proposed~\cite{sdp_rob_local}, we are unaware of any existing SDP tools to certify $\ell_2$-robustness. The $\ell_2$-robustness certification is an active topic~\cite{bcp,gloro}.

\cite{convex_barrier} tried to unify efficient robustness certification works, which the authors claimed is a convex relaxation barrier. However, they only studied LP relaxation methods and were unable to include the SDP certification method~\cite{sdp_rob_local}. As we have discussed, SDP is strictly more powerful than LP. Our framework can be viewed as complementing~\cite{convex_barrier} towards a tighter convex relaxation barrier.

\section{Related Work}
Using SDP to verify programs is not new to the verification community.~\cite{semialgebraic-verify} applied SDP to verify the invariance and termination of semialgebraic programs.~\cite{sdp-prob} used SDP to verify the termination of probabilistic programs. From the verification point of view, our framework studies how to  abstract the neural-network properties into SDPs. In particular, we use symbolic domains and quadratic relations to achieve this abstraction. Meanwhile, semidefinite programming as a unique subject has also developed remarkably over the past decades. In particular, many SDP-based algorithms for intractable problems are believed optimal within polynomial time\cite{opt_inf2,ugc_cut,UGC_inf1_opt}. We provide a comprehensive discussion on many aspects of the SDP relaxation, which can in return bring more perspectives to other SDP-enabled verification works.

\section{Conclusion}
In this work, we present a new program reasoning framework for DNN verification. We have demonstrated its power on several network verification tasks, and also discussed many aspects of this paradigm. We believe that this framework can bring new representations and perspectives of the verification tasks and more tools to study them. 
%
%
%
\bibliographystyle{splncs04}
\bibliography{paper}

\appendix
\section{Quadratic Encodings for $\ell_p$-norm Constraints}\label{sec:ell-p}
In the canonical finite-dimensional vector space $\R^n$, the most commonly considered $\ell_p$-norms in practice are the $\ell_2$ and $\ell_\infty$-norms. $\ell_1$-attacks are also considered in some literatures~\cite{jordan2021exactly}. We have shown how to encode $\ell_2$ and $\ell_\infty$-norm constraints in the main text. 

The $\ell_1$-norm constraint can be encoded via Hölder's inequality:
\[\norm{x}_1\leq \epsilon \Leftrightarrow \inprod{x, y}\leq \epsilon , \norm{y}_\infty\leq 1.\]
Therefore, one can introduce a new variable $y$ to encode the $\ell_1$-norm constraint.

For more general $\ell_p$-norms, where $p\geq 1$ and is rational, we can use the following encoding, as described in~\cite{modern_co}.

First, let $q$ be the Hölder conjugate of $p$, i.e., $\frac{1}{p} + \frac{1}{q} = 1$. For example, the dual norm of $\ell_2$-norm is $\ell_2$ because $\frac{1}{2} + \frac{1}{2} = 1$, and also $1$ and $\infty$ are dual to each other. One can verify that $\norm{x}_p\leq \epsilon$ is equivalent to the following conditions:
\begin{equation}\label{eq:ell-p}
    y_i\geq 0, |x_i|\leq \epsilon^{1/q}y_i^{1/p}, \sum_{i=1}^n y_i\leq \epsilon.
\end{equation}
This is because by definition, $\norm{x}_p\leq \epsilon$ means that $\sum_{i=1}^n |x_i|^p\leq \epsilon^p$. One can then think of $y_i$ as $|x_i|^p\epsilon^{1-p}$.

The linear equations can be easily encoded by quadratic relations. It remains to encode the inequality $x_i\leq \epsilon^{1/q}y_i^{1/p}$ in quadratic relations. This is easy because $\epsilon^{1/q}$ is a number, so we only need to consider how to express $x_i\leq y_i^{q}$.

Because $q$ is a rational number, the above inequality is equivalent to $x_i^{n}\leq y_i^m$, for integers $n$ and $m$. If $n, m \geq 3$, we can introduce additional variables and inequalities to decrease the degree. For example, $x^3\leq y$ can be encoded as $xz\leq y$ and $z=x^2$.

\section{Quadratic encodings for other activations}\label{sec:other-act}
Let $\act$ be the activation function, and $z=\act(x)$.

If we consider the data-independent analysis and the slope-restricted interpretation of the activation, we only need to know the derivative bounds of the activation function. Most activation functions such as $\relu$, sigmoid functions are slope-restricted, and the bounds can be easily found. Suppose the lower and upper bounds are $a$ and $b$, then we can use $(\Delta z - a\Delta x)(\Delta z - b\Delta x)\leq 0$ to capture this interpretation.

The complicated case is the exact encoding of the activation functions. One can easily see that quadratic relations cannot exactly encode non-algebraic functions. If the activation contains $e^x$ in its irreducible representation, for example, the hyperbolic tangent function $\tanh(x)=\frac{e^x-e^{-x}}{e^x+e^{-x}}$, we can only approximate them. One can consider segment the input space into several regions and then use two linear piecewise functions to upper and lower bound the non-algebraic function. Because piecewise linear functions can approximate any continuous function within any precision on a compact space, we can bound any continuous activation function up to arbitrary precision. It remains to approximate the piecewise linear function. 

In~\cite{relu-express}, the authors showed that $\relu$ network can exactly encode any piecewise linear functions. Therefore, one can use the $\relu$ quadratic encoding as the block to construct any piecewise linear function. To illustrate this process, we use $\relu_\theta$ as an example. $\relu_\theta$ is used in~\cite{huang2021local} to train locally Lipschitz network. $\relu_\theta$ is defined as
\[ \relu_\theta(x) = \left\{ \begin{array}{ll}
            \theta, & \mbox{$x \geq \theta$}\\
            x, & \mbox{$0\leq x < \theta$}\\
            0, & \mbox{$x < 0$}\end{array} \right.
        \]
where $\theta>0$.

We can use a similar squashable function gadget in~\cite{IUA} to construct $\relu_\theta$ from $\relu$. One can easily check that $\relu_\theta(x) = \relu(\theta-\relu(\theta-x))$.

Because $z = \relu_\theta(x)$, we want to constrain $z$ and $x$ with quadratic relations. We can introduce another variable $y = \relu(\theta-x)$. From $\relu$'s quadratic encoding, we have 
$y\geq \theta-x$, $y\geq 0$ and $(y-(\theta-x))y\leq 0$.

Similarly, we have $z\geq \theta-y$, $z\geq 0$ and $(z-(\theta-y))z\leq 0$.

With these six quadratic inequalities, we can encode the computation $z = \relu_\theta(x)$.

\section{Shor's Relaxation Scheme}\label{sec:shor}
We provide a detailed introduction of Shor's relaxation scheme as in~\cite{modern_co}. The quadratic program is:
\begin{align*}
\begin{split}
    \min \;\;\;&f_0(x) = x^TA_0x+2b_0^Tx+c_0\\ 
    s.t.\;\;\;\;  & f_i(x) = x^TA_ix+2b_i^Tx+c_i \leq 0,\;\forall i\in[m]
\end{split}
\end{align*}

\paragraph{Dual form SDP relaxation}One can introduce $m$ nonnegative variables $\lambda_i\geq 0$ as ``weights'' for the constraints. These non-negative variables are known as the Lagrangian dual variables. One can add the weighted constraint to the objective, and obtain:
\[f_\lambda(x) = f_0(x)+\sum_{i=1}^m\lambda_if_i(x)=x^TA(\lambda)x+2b^T(\lambda)x+c(\lambda),\]
where
\begin{align*}
    A(\lambda) &= A_0 + \sum_{i=1}^m \lambda_iA_i,\\ 
    b(\lambda) &= b_0 + \sum_{i=1}^m \lambda_ib_i,\\ 
    c(\lambda) &= c_0 + \sum_{i=1}^m \lambda_ic_i.
\end{align*}
Because $f_i(x)\leq 0$ and $\lambda_i\geq 0$, by construction, $\inf_{\lambda}f_\lambda(x)$ lower bounds the optimum of~\cref{eq:quad-prog}, and thus a relaxation. The interesting part of this weighted optimization problem is that this can be represented as an SDP. The minimization of of $f_\lambda(x)$ is to maximize $\obju$ such that $f_\lambda(x) - \obju\geq 0$
. This problem has an SDP representation:
\[
\max_{\obju,\dualv}\Big\{\obju:\begin{pmatrix}
c(\lambda)-\obju & \;\;\;\;b(\lambda)^T \\
b(\lambda) & \;\;\;\;A(\lambda)
\end{pmatrix}\succeq 0, \dualv_i\geq 0\Big\}.
\]
This is Shor's semidefinite relaxation of the quadratic program in the dual form.

\paragraph{Primal form SDP relaxation}We have introduced the primal form SDP relaxation in~\cref{sec:overview}. Now we can discuss why the primal SDP relaxation can be viewed as the natural continuous relaxation for some combinatorial problems. Essentially this is because the discrete constraint quantified by a quadratic relation is relaxed to a continuous constraint by adding more dimensions.  We can use the MAXCUT problem as an example.

The MAXCUT problem is defined as $\max_{x\in\{-1,1\}^n} x^TLx$, where $L\in\R^{n\times n}$ is the Laplacian matrix of a graph.

The SDP relaxation for this problem is
\begin{align*}
    &\inprod{L,X}_F\\
    s.t.\;\;&X\succeq 0, X_{ii} = 1.
\end{align*}
Because $X$ is PSD, $X = MM^T$ for some matrix $M\in\R^{n\times d}$, where $d\in\Z_+$. Let $M_i$ be the $i$-th row vector of $M$. $X_{ij} = \inprod{M_i,M_j}$, and $X_{ii}=1$ means $\inprod{M_i, M_i} = 1$. As a result, $\inprod{L,X}_F = \sum_{i,j} L_{ij}X_{ij} = \sum_{i,j} L_{ij}\inprod{M_i, M_j}$.
In contrast, in the quadratic program, the variable to $L_{ij}$ is $x_ix_j$, the product of two scalars. 
If $d=1$ in the SDP relaxation, $M$ is a column vector, and $X$ is a rank-1 matrix. In this case, the SDP coincides with the combinatorial problem, because the inner product degenerates to the multiplication of two scalars. Hence, the SDP relaxation can be viewed as a continuous relaxation of a discrete problem.

\section{An Example for Recursion Reasoning}\label{sec:rec-prob}
Let us consider a classical geometric distribution example. Given a fair coin, suppose we want to flip the coin until we see heads for the first time, what is the expected number of trials?

There are two approaches to this problem. We can define a series and then find the limit: We need $1$ trial with probability $1/2$; $2$ trials with probability $(1/2)^2$, etc. So the expected number of trials is 
\[
\sum_{i=1}^\infty \frac{i}{2^i}.
\]
The limit of this series is $2$.

We can build a recursion: If we flip the coin, we either get a head with probability $1/2$, or get a tail. In the tail case, we are essentially the same as in the beginning. This actually forms a Markov chain or a probabilistic finite state machine. Therefore, $T = 1+0*(1/2)+T*(1/2)$. Solving this formula gives $T = 2$. Similar reasoning is used for calculating the absorbing probability of Markov chains, which is a recurrent system.

For the second approach, we reason the recursion directly by assigning symbols to the state and constructing a symbolic formula. Therefore, we can reason the limiting behavior directly. We use this example to demonstrate how universal and powerful symbolic reasoning is.




\section{Proof of~\cref{thm:weighted-avg}}\label{sec:proof}
\begin{proof}
From Rayleigh quotient, for any symmetric matrix $M$, $\eig_{\max}(M) = \max x^T M x$, where $x^Tx = 1$. This is a quadratic program, and its Shor's relaxation is:
\begin{align*}
\begin{split}
   &\max  \inprod{M,X}_F\\
    s.t.\; & X \succeq 0, \trace(IX)=1.
\end{split}
\end{align*}
The dual program is:
\begin{align}\label{eq:max-eigval}
\begin{split}
   &\min  t\\
    s.t.\; & tI-M\succeq 0.
\end{split}
\end{align}
Notice that by S-lemma, this relaxation is exact.

Let us consider the following semidefinite program:
\begin{align*}
\begin{split}
   &\max  \inprod{M,X}_F\\
    s.t.\; & X \succeq 0, X_{ii}=1, i\in [n];
\end{split}
\end{align*}
where $M$ is a symmetric matrix. From Slater's condition, the strong duality holds, and this program has the same value as its dual formulation:
\begin{align*}
\begin{split}
   &\min  \sum_{i=1}^{n} y_i\\
    s.t.&\;\;  \ACTL(y)-M\succeq 0;
\end{split}
\end{align*}
for $y_i\geq 0$. Because we want to use the $\lambda_{\max}$-SDP, we cannot translate $X_{ii}=1$ to $\trace(IX)=1$ directly. However, we can consider a \emph{corrected average} approach to do this translation. 

Let $t = \frac{\sum y_i}{n}$. Therefore, $y = te_{n} - h$ for some $\sum_{i=1}^{n} h_i = 0$, in which we can consider $h$ as a correction term for the average. Then the objective becomes $\sum y_i = n t$, and $\ACTL(y) = \ACTL(te_{n}-h) = tI - \ACTL(u)$. So the dual program can be rewritten as
\begin{align*}
\begin{split}
   &\min\;  nt\\
    s.t.\;\;  tI &- (M+\ACTL(h))\succeq 0;
\end{split}
\end{align*}
where $\sum h_i = 0$. From~\cref{eq:max-eigval}, this is equivalent to $n\min_h \eig_{\max}(M+\ACTL(h))$. Solving the SDP amounts to finding the optimal \emph{correcting vector} $u$ that minimizes the maximum eigenvalue of $M+\ACTL(h)$. 

This concludes the proof of~\cref{thm:weighted-avg}.
\end{proof}

\cite{certified_def} used a similar result to train the network with the SDP constraint because $\eig_{\max}$ is a differentiable function and can provide an estimation of the FGL immediately. Their bound is
\[\min_{c\in\R_+^{n}}(\sum_{i=1}^{n} c_i + n\eig_{\max}^+(M-\ACTL(c)),\] 
where $\eig_{\max}^+(x) = \max(\eig_{\max}(x),0)$.

However, our bound is \[\min_h n\eig_{\max}(M+\ACTL(h)),\] where $\sum h_i=0$, and for any symmetric $M$.

Our bound is tighter. Consider
$M = \begin{pmatrix}
1 & 0\\
0& -1
\end{pmatrix}$. We can choose the weighted average vector $h = (-1,1)$, so $M+\ACTL(h)$ is the zero matrix, and $\eig_{\max}(M+\ACTL(h)) = 0$. While for any $c\in \R_+^2$, $\sum_{i=1}^{n} c_i + 2\eig_{\max}^+(M-\ACTL(c)) > 0$.



\end{document}